\pdfoutput=1

\documentclass[11pt]{article}

\usepackage[preprint]{acl}

\usepackage{times}
\usepackage{latexsym}
\usepackage{lipsum}
\usepackage{amsmath}
\usepackage{tcolorbox}
\usepackage{placeins} 
\usepackage{float}
\usepackage{multirow}

\usepackage{booktabs}

\usepackage[T1]{fontenc}

\usepackage[utf8]{inputenc}

\usepackage{microtype}

\usepackage{inconsolata}

\usepackage{graphicx}

\usepackage{tcolorbox}

\newtcolorbox{codebox}{colback=gray!20, colframe=gray!80, boxrule=0.5pt, sharp corners, enhanced, breakable}

\newcommand{\subsec}[1]{\noindent\textbf{#1}\quad}
\newcommand{\fv}{\( \mathcal{FV} \)}
\newcommand{\cv}{\( \mathcal{CV} \)}

\title{Analogical Reasoning Inside Large Language Models:\protect\\Concept Vectors and the Limits of Abstraction}

\author{Gustaw Opiełka, Hannes Rosenbusch, Claire E. Stevenson \\
  Department of Psychological Methods, University of Amsterdam \\
  \texttt{g.j.opielka@uva.nl} 
  }

\begin{document}
\maketitle
\begin{abstract}
Analogical reasoning relies on conceptual abstractions, but it is unclear whether Large Language Models (LLMs) harbor such internal representations. We explore distilled representations from LLM activations and find that function vectors (\fv s; \citealp{toddFunctionVectorsLarge2024})—compact representations for in-context learning (ICL) tasks—are not invariant to simple input changes (e.g., open-ended vs. multiple-choice), suggesting they capture more than pure concepts. Using representational similarity analysis (RSA), we localize a small set of attention heads that encode invariant concept vectors (\cv s) for verbal concepts like \textit{antonym}. These \cv s function as feature detectors that operate independently of the final output—meaning that a model may form a correct internal representation yet still produce an incorrect output. Furthermore, \cv s can be used to causally guide model behaviour. However, for more abstract concepts like \textit{previous} and \textit{next}, we do not observe invariant linear representations, a finding we link to generalizability issues LLMs display within these domains.
\end{abstract}

\begin{figure}[t]
  \includegraphics[width=\columnwidth]{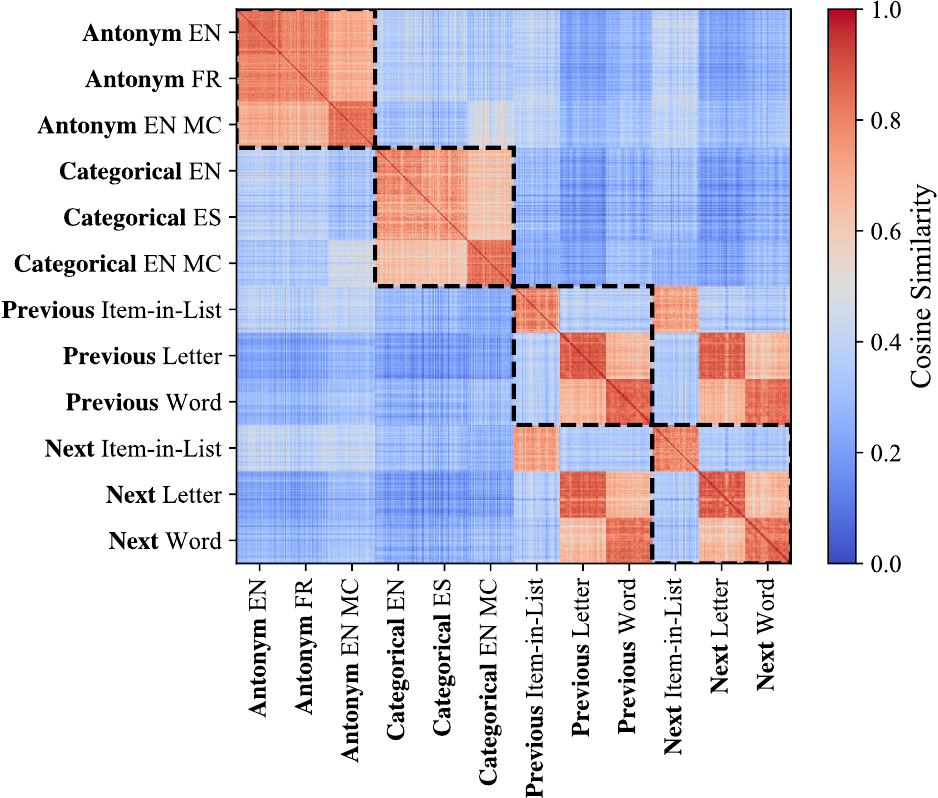}
  \caption{Pairwise similarity matrix of \cv s extracted from Llama-3.1 70B across 600 ICL prompts covering various concepts and low-level presentations. \cv s remain invariant for the verbal concepts \textit{antonym} and \textit{category}, but show no stable representation of abstract concepts like \textit{previous} or \textit{next}. Instead, these tasks exhibit order-based representations tied to known lists (e.g., alphabets, weekdays) or low-level clustering based on presentation format (words vs. letters).}
  \label{fig:main}
\end{figure}

\section{Introduction}
"Analogies are functions of the mind" \cite[][p.10]{hillLearningMakeAnalogies2019a}. People use analogies to flexibly map previous knowledge to novel domains \citep{hofstaderGodelEscherBach1979, mitchellArtificialIntelligenceGuide2020}. For example, if you are just beginning to learn about analogical reasoning, envisioning a “bridge” that connects new information to concepts you already understand can be very helpful. In essence, successful analogy-making depends on our ability to extract and apply conceptual abstractions—such as “bridge” or “connection”—from seemingly unrelated situations.  While behavioral evidence suggests that analogical reasoning have emerged in LLMs \citep{brownLanguageModelsAre2020, webbEmergentAnalogicalReasoning2023}, it remains unclear if and how LLMs represent these relational concepts internally.

What does it mean for a neural system to represent abstract concepts? We formalize abstraction as \textit{conceptual invariance}.

Consider a high-level concept \( \mathcal{C} \) (e.g., "antonym"). A neural network \( f \) flexibly represents \( \mathcal{C} \) if it encodes the same abstract representation regardless of variations in its low-level inputs. Let \( X \) denote the space of all inputs encoding \( \mathcal{C} \) and \( \mathcal{T} \) a group of transformations on \( X \) (e.g., changes in language, format, or modality) that preserve the concept’s meaning. Then, \( f \) satisfies conceptual invariance if

\[
f(t(x)) = f(x), \quad \forall t \in \mathcal{T}.
\]

This ensures that the network’s encoding of \( C \) reflects its essence rather than superficial characteristics of low-level input. This is analogous to how object representations in convolutional neural networks are translation-invariant \citep{lecunGradientbasedLearningApplied1998}.

\subsec{Previous Work} Previous work identified \textit{Function Vectors} (\fv s; \citealp{toddFunctionVectorsLarge2024}; \citealp{hendelInContextLearningCreates2023}), a compact vector representation of an ICL task \citep{brownLanguageModelsAre2020}. The representation is encoded by a universal set of attention heads (high overlap of heads across different tasks), and can be transplanted into the model internals to causally guide its behavior (even zero-shot - e.g. transplanting an antonym \fv{} to a prompt 'fast: ' induces the network output 'slow'). Attention heads composing the \fv{} were found using activation patching a popular mechanistic interpretability technique for localizing information in neural networks (\citealp{heimersheimHowUseInterpret2024}; Details in Sec. \ref{AP}).

\subsec{Summary of contributions} 
We investigate whether conceptual invariance holds for \fv s and find they are not invariant to low-level changes (e.g., switching the ICL format from open-ended to multiple-choice; Sec. \ref{fv_no_invariance}). Instead \fv s encode dense, detailed information that goes beyond the latent conceptual content we were targeting (Sec. \ref{fv_are_rich}). Based on additional checks we conclude that activation patching itself may be responsible for this shortcoming, as it appears to overlook the true latent representations (Sec. \ref{ap_issue}).

We then use representational similarity analysis (RSA; \ref{RSA}) to localize latent abstract information in transformer internals. For verbal concepts, we find a set of attention heads emerging in early-to-mid layers (Sec. \ref{cv_yes_invariance}). By summing their outputs we form the \cv. We find that the extent of conceptual invariance grows with number of training examples in the ICL prompts. Interestingly, we find that \cv s can carry the correct conceptual representations while the model produces incorrect answers (Sec. \ref{cv_feature}). We then ask whether the \cv s causally influence behaviour (Sec. \ref{cv_causal}). We find that while being much weaker at zero-shot interventions, with enough context in the prompt, \cv s influence model output and do so in a more portable manner than \fv s. 

Finally, we use \cv s to demonstrate that our LLMs did not develop representations of abstract concepts of 'Previous' and 'Next' (Figure \ref{fig:main}), which hinders their generalization to letter-string tasks (Sec. \ref{generalization}). We use our findings to inform the discussion of analogical reasoning capabilities in LLMs through the lens of internal representations.

\begin{table*}[t]
  \centering
  \footnotesize 
  \setlength{\tabcolsep}{4pt} 
  \renewcommand{\arraystretch}{0.9} 
  \begin{tabular}{lccccc}
    \toprule
    \textbf{Concept} & \textbf{Dataset} & \textbf{Question Type} & \textbf{Response Type} & \textbf{Info Source} & \textbf{Lang} \\
    \midrule
    \multirow{3}{*}{Translation} & English to French    & open   & word   & not in prompt & FR \\
                                 & German to Spanish    & open   & word   & not in prompt & ES \\
                                 & English to French-MC & MC     & letter & in prompt     & -  \\
    \midrule
    \multirow{3}{*}{Antonym}     & Antonym EN           & open   & word   & not in prompt & EN \\
                                 & Antonym FR           & open   & word   & not in prompt & FR \\
                                 & Antonym MC           & MC     & letter & in prompt     & -  \\
    \midrule
    \multirow{3}{*}{Categorical} & Categorical EN       & open   & word   & not in prompt & EN \\
                                 & Categorical ES       & open   & word   & not in prompt & ES \\
                                 & Categorical MC       & MC     & letter & in prompt     & -  \\
    \midrule
    \multirow{3}{*}{Previous}    & Prev Item-in-List    & open   & mixed  & not in prompt & -  \\
                                 & Prev Abstract-Letter & open   & letter & in prompt     & -  \\
                                 & Prev Abstract-Word   & open   & word   & in prompt     & EN \\
    \midrule
    \multirow{3}{*}{Next}        & Next Item-in-List    & open   & mixed  & not in prompt & -  \\
                                 & Next Abstract-Letter & open   & letter & in prompt     & -  \\
                                 & Next Abstract-Word   & open   & word   & in prompt     & EN \\
    \bottomrule
  \end{tabular}
  \caption{Task Information Table}
  \label{tab:task_info}
\end{table*}

\section{Materials and Methods}

\subsection{Models}
We investigate the LLama 3.1 model family \citep{grattafioriLlama3Herd2024}, specifically on the 8 and 70 billion parameter variants. 

Llamas are autoregressive, residual-based transformers. The models, \( f \) internally comprise of \( \mathcal{L} \) layers. Each layer is composed of a multi-layer perceptron (MLP) and \( J \) attention heads \( a_{\ell{j}}\) which together produce the vector representation of the last token, \(\mathbf{h}_{\ell} = \mathbf{h}_{\ell-1} +  \text{MLP}_\ell + \sum_{j\in{J}} a_{\ell{j}} \) \citep{elhage2021mathematical}. In all our experiments we focus on the representations extracted from the last token position.

\subsection{Task Formulation}
For every dataset \( d \in D \) in our collection, we define a set \( P_d \) containing in-context prompts \( p_d^i \in P_d \).

Each prompt \( p_d^i \) is a token sequence that includes \( N \) input-output exemplar pairs \((x, y)\), all illustrating the same underlying concept \( \mathcal{C} \) and its corresponding mapping from \( x \) to \( y \). Additionally, each prompt provides a query input \( x_q^i \) linked to a target response \( y_q^i \). \( y_q^i \) is not shown to the model and we consider that the model performs correctly on \( p_d^i \) if its predicted token matches \( y_q^i \) (or the first token of \( y_q^i \) for multi-token words). 

\subsection{Verbal Concepts}

\subsec{Translation} We use English-to-French and German-to-Spanish tasks.

\subsec{Antonym} We source antonym word pairs from \citet{toddFunctionVectorsLarge2024}. E.g.,: \texttt{Big} $\rightarrow$ \texttt{Small}.

\subsec{Categorical} We generate 1000 pairs using OpenAI's GPT-4o. E.g.,: \texttt{Table} $\rightarrow$ \texttt{Furniture}.

\vspace{5pt}
\subsec{Low-level transformations} We test verbal concepts in three low-level presentations - Open-ended in English, Open-ended in a different language, and Multiple-Choice (MC) in English.

\subsection{Abstract Concepts} \label{abstract_concepts}

We investigate two abstract concepts, \textbf{Previous} and \textbf{Next}, capturing whether an entity comes before or after another entity. We test these concepts using three different low-level presentations:

\subsec{Item in List} Our pairs are made up of days of the week, months of the year, letters of the alphabet, and number pairs (both numeric and text form). 
\vspace{0.5pt}
\noindent Some examples for Next-Item in List: 
\texttt{Monday} $\rightarrow$ \texttt{Tuesday}, \texttt{December} $\rightarrow$ \texttt{January}, \texttt{a} $\rightarrow$ \texttt{b}, \texttt{seven} $\rightarrow$ \texttt{eight}. 

\noindent And for Previous-Item in List:
\texttt{Tuesday} $\rightarrow$ \texttt{Monday}, \texttt{January} $\rightarrow$ \texttt{December}, \texttt{a} $\rightarrow$ \texttt{z}, \texttt{eight} $\rightarrow$ \texttt{seven}.

\subsec{Abstract Previous/Next Task} We evaluate tasks where a sequence contains one indicator element, one target element, m distractors sharing the target’s features, and n positional elements that do not. The target always appears either before (Previous) or after (Next) the indicator. We test two variants—using either English words or letters (a, b, c, d)—with one-token elements. Below we show examples for \(m = 3\), \(n = 3\) with indicator elements being "\texttt{\textbf{*}}" and positional "\texttt{\textbf{.}}". The target elements are "\texttt{\textbf{c}}" and "\texttt{\textbf{letter}}".

{
  \centering
  \begin{tcolorbox}[
    width=0.7\linewidth,
    colback=white,
    colframe=black,
    sharp corners,
    boxrule=0.3pt,
    boxsep=2pt,
    left=2pt,
    right=2pt,
    top=2pt,
    bottom=2pt,
    before skip=5pt,
    after skip=5pt
  ]
    {\footnotesize\textbf{Previous-Letter Example:}}\\[2pt]
    {\small\ttfamily
      \begin{tabular}{@{}l l@{}}
        Q: . a c . * b . d   & A: \textcolor[HTML]{a53e3a}{c} \\
        Q: c a * . . d b .   & A: \textcolor[HTML]{a53e3a}{a} \\
        Q: b a d c . . *      & A:
      \end{tabular}
    }
  \end{tcolorbox}
}

{
  \centering
  \begin{tcolorbox}[
    width=0.8\linewidth,
    colback=white,
    colframe=black,
    sharp corners,
    boxrule=0.3pt,
    boxsep=2pt,
    left=2pt,
    right=2pt,
    top=2pt,
    bottom=2pt,
    before skip=5pt,
    after skip=5pt
  ]
    {\footnotesize\textbf{Next-Word Example:}}\\[2pt]
    {\small\ttfamily
        Q: . big mask . * control . house  \\
        A: \textcolor[HTML]{a53e3a}{control} \\
        Q: star code * . . dense light . \\ 
        A: \textcolor[HTML]{a53e3a}{dense} \\
        Q: ball might poland  * . letter .  \\ 
        A:
    }
  \end{tcolorbox}
}

\subsection{Task Attributes}
Our tasks have high-level (concepts) and low-level attributes: \textbf{Question Type} - ICL prompt in either open-ended or multiple-choice (MC) format; \textbf{Response Type} - whether the expected response is a word, letter, or a mix of both; \textbf{Information Source} - whether the expected response is located somewhere in the prompt (e.g., MC items), or needs to be generated (e.g., open-; \textbf{Language}-the language of the expected response.

\subsection{Activation Patching} \label{AP}

Activation patching replaces specific activations with cached ones from a \textit{clean} run to assess their impact on the model’s output. The cached activations are then inserted into selected model components in a \textit{corrupted} run, where the systematic relationships in the prompt are disrupted. For example, in an antonym ICL task, consider a \textit{clean prompt}:

{\small
\begin{verbatim}
Hot -> Cold : Big -> Small : Clean -> ?
\end{verbatim}
}
\noindent and a \textit{corrupted prompt}:
{\small
\begin{verbatim}
House -> Cold : Eagle -> Small : Clean -> ?
\end{verbatim}
}

To localize attention heads carrying task-relevant information we compute the \emph{causal indirect effect} (CIE) for each attention head \( a_{\ell{j}} \) as the difference between the probability of predicting the expected answer \( y \) when processing the corrupted prompt \( \tilde{p} \) with and without the transplanted mean activation \( \bar{a}_{\ell{j}} \) from clean runs:
\[
\text{CIE}\big(a_{\ell{j}}\big) = f\Big(\tilde{p} \mid a_{\ell{j}} := \bar{a}_{\ell{j}}\Big)[y] - f\big(\tilde{p}\big)[y].
\]

We then compute the \emph{average indirect effect} (AIE) over a collection \( \mathcal{D} \) of 10 datasets from \citet{toddFunctionVectorsLarge2024}:
\[
\text{AIE}(a_{\ell{j}}) = \frac{1}{|\mathcal{D}|} \sum_{d \in \mathcal{D}} \frac{1}{|\tilde{\mathcal{P}}_d|} \sum_{\tilde{p}_i \in \tilde{\mathcal{P}}_d} \text{CIE}\big(a_{\ell{j}}\big),
\]
where \( \tilde{\mathcal{P}}_d \) denotes the set of corrupted prompts for dataset \( d \).

\subsection{Function Vectors} \label{FVs}
A function vector for a specific dataset (\(\mathcal{FV}_{d}\)) is computed as the sum of the mean activations over all clean prompts from the dataset from a set \( \mathcal{A}_{\mathcal{FV}} \) of top \( N \) attention heads having the highest AIE values:
\[
\mathcal{FV} = \sum_{a_{\ell{j}} \in \mathcal{A}} \bar{a}_{\ell{j}}.
\]
Following the implementation in \citet{toddFunctionVectorsLarge2024}, we set \( N = 20 \) for the 8B model and \( N = 100 \) for the 70B model.

\subsection{Representational Similarity Analysis} \label{RSA}
To distill conceptual information from LLMs during ICL, we employ representational similarity analysis (RSA)—a technique invented for cognitive neuroscience \citep{kriegeskorteRepresentationalSimilarityAnalysis2008a}. In our work, RSA is used to assess the alignment between LLM representations and task attributes.

For each \( a_{\ell{j}} \) we compute representational similarity matrices (RSMs) of the form:
\[
\text{RSM} =
\begin{bmatrix}
    1 & \cdots & \theta(v_1, v_N) \\
    \vdots & \ddots & \vdots \\
    \theta(v_N, v_1) & \cdots & 1
\end{bmatrix}
\]
where \( v_i \) denotes the output extracted from \( a_{\ell{j}} \) for the \( i \)th prompt \( p_i \in P_N \), and \( \theta(\cdot,\cdot) \) is a similarity function.

Additionally, for each task attribute \( q \) (i.e., \texttt{concept}, \texttt{info\_source}, \texttt{lang}, \texttt{response\_type}, \texttt{task\_type}), we construct \( N \times N \) binary design matrix \( \text{DM}_q \), where each entry is set to 1 if the corresponding pair of prompts share the same attribute value, and 0 otherwise.

We then quantify the alignment between the lower-triangular portions of the RSM and \( \text{DM}_q \) using the non-parametric Spearman's rank correlation coefficient. This alignment for \( a_{\ell{j}} \) is denoted by \( \Phi_{\ell{j}}^q \). When referring to \( \Phi^{\texttt{concept}} \) we mean the alignment between model activations and the subset of datasets containing \textit{verbal} concepts only, unless stated otherwise.

\subsection{Concept Vectors}
Analogous to \fv s (Sec. \ref{FVs}), the (\(\mathcal{CV}_{d}\))'s   are constructed by summing the mean activations from a set of top-ranking attention heads. In this case, we sum the top 3 attention heads with the highest \( \Phi^{\texttt{concept}} \) scores, forming a set \( \mathcal{A}_{\mathcal{CV}} \), for both model sizes.

\section{Do \fv s create an invariant representation of latent concepts?}

We start our search for invariant conceptual representations using methods that rely on activation patching. We show that \fv s carry more than purely relational information, and that diversifying the datasets does not help localize the attention heads carrying latent information.

\begin{figure}[t]
  \includegraphics[width=\columnwidth]{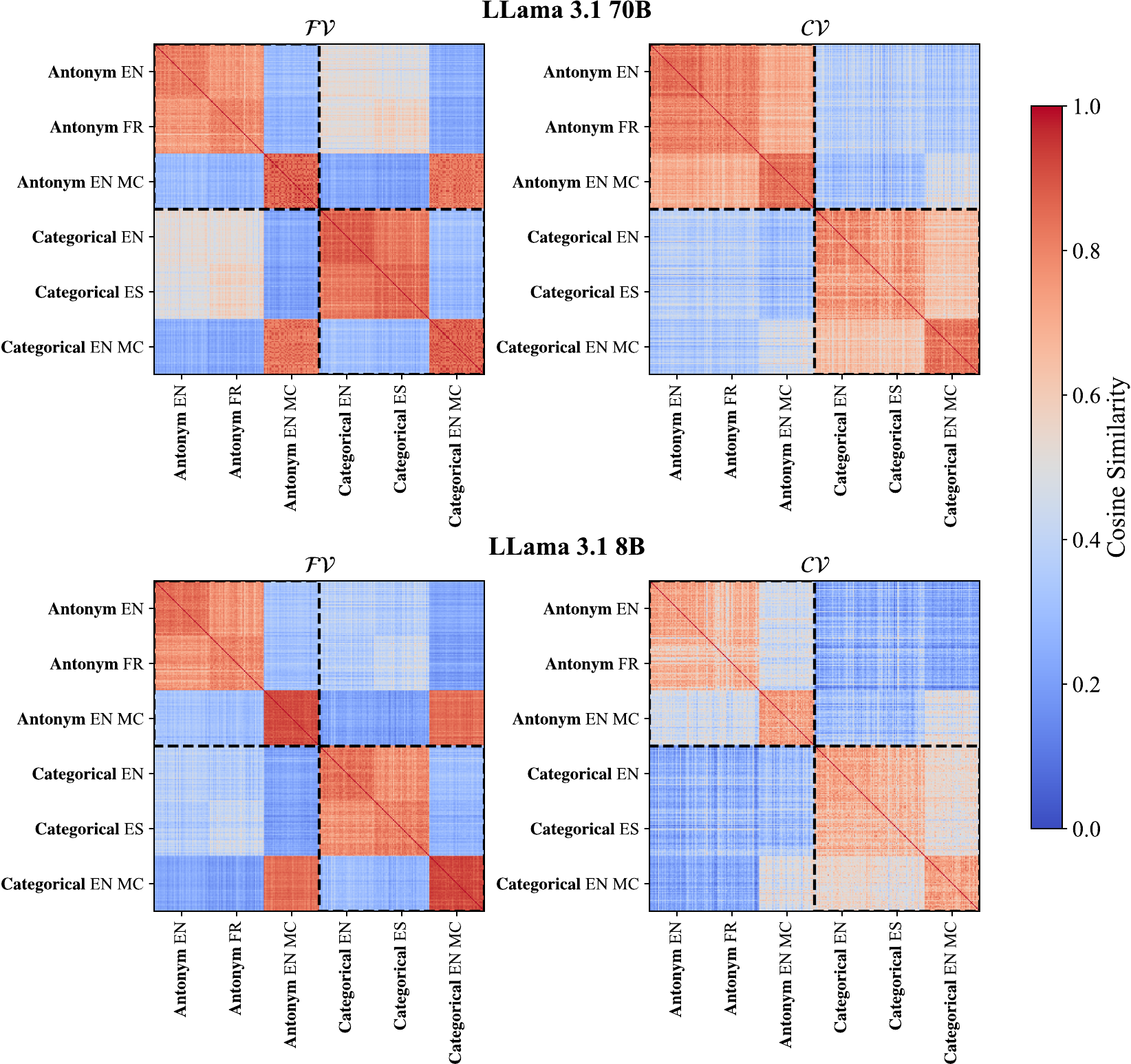}
  \caption{Representational similarity matrices for antonym and categorical concepts each tested with three low-level transformations. The upper-left and lower-right quadrants (outlined with the dashed lines) contain pairwise similarity scores for prompts coming from the same concept. \cv s encode the concept in a more invariant manner than \fv s.  }
  \label{fig:fv_vs_cv}
\end{figure}

\subsection{\fv s are not invariant to low-level transformations} \label{fv_no_invariance}

We extract \fv s for all datasets and items (Table \ref{tab:task_info}). That is, for prompt \(i  \in N\) prompts \(\mathcal{FV}_i = \sum a_{\ell{j}}^i\), where \(a_{\ell{j}} \in A_{\mathcal{FV}}\). Each dataset has 50 prompts, each consisting of a 5-shot ICL task.

As we see in Figure \ref{fig:fv_vs_cv} \fv{} representations cluster within the concepts in both languages in open-ended question formats, but the clustering disappears for multiple-choice prompts, where all items cluster together, despite encompassing multiple concepts (e.g., antonym and categorical MC items show high similarity - they are represented using a subspace that is orthogonal to open-ended items). This shows that \fv{} representations are contextual rather than conceptually invariant.

\subsection{\fv s encode multiple task attributes} \label{fv_are_rich}
This leads us to the question \textit{what} information \fv s encode, if not purely the concepts? We answer this by investigating how much each task attribute explains the activation spaces of each attention head in \( A_{\mathcal{FV}} \). 

\begin{figure}[t]
  \centering
  \includegraphics[width=0.8\columnwidth]{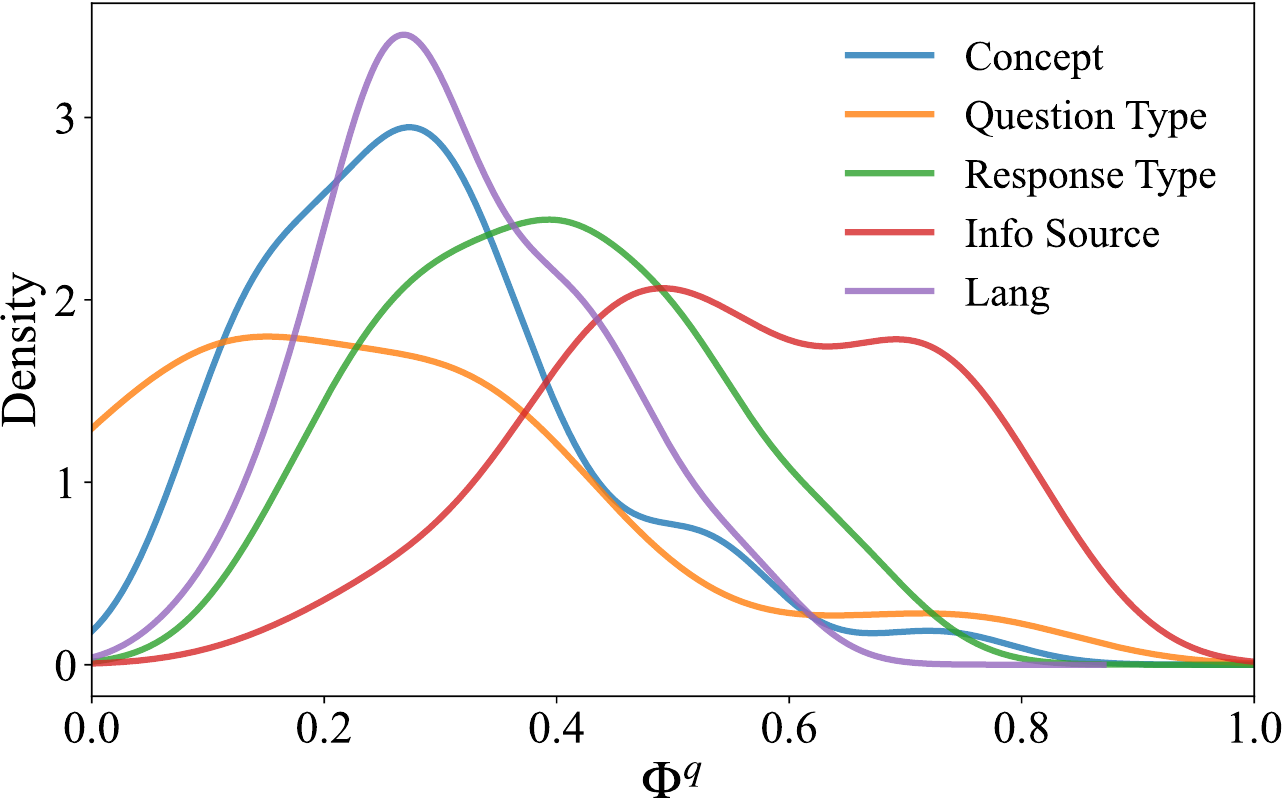}
  \caption{Density plot displaying the information-rich make-up of 100 attention heads in LLama 70B comprising its \fv.}
  \label{fig:density}
\end{figure}

Figure \ref{fig:density} displays density plots for all \( \Phi_{\ell{j}}^q \). These plots reveal that each task attribute is represented to some extent within the \fv s, with \texttt{task\_type} exhibiting the highest density. This indicates that the attention heads forming the \fv s are particularly sensitive to whether the language model is tasked with extracting information from the input prompt or generating a novel token. This sensitivity aligns with the RSM shown in Figure \ref{fig:fv_vs_cv}—multiple-choice items form distinct clusters because they are extractive (in contrast to open-ended items) and have a different response type (four possible letters versus words). Importantly, while relational information is present, it does not play a crucial role in shaping the \fv s, confirming that \fv s are not invariant representations of latent concepts.

\begin{figure}[t]
  \centering
  \includegraphics[width=0.8\columnwidth]{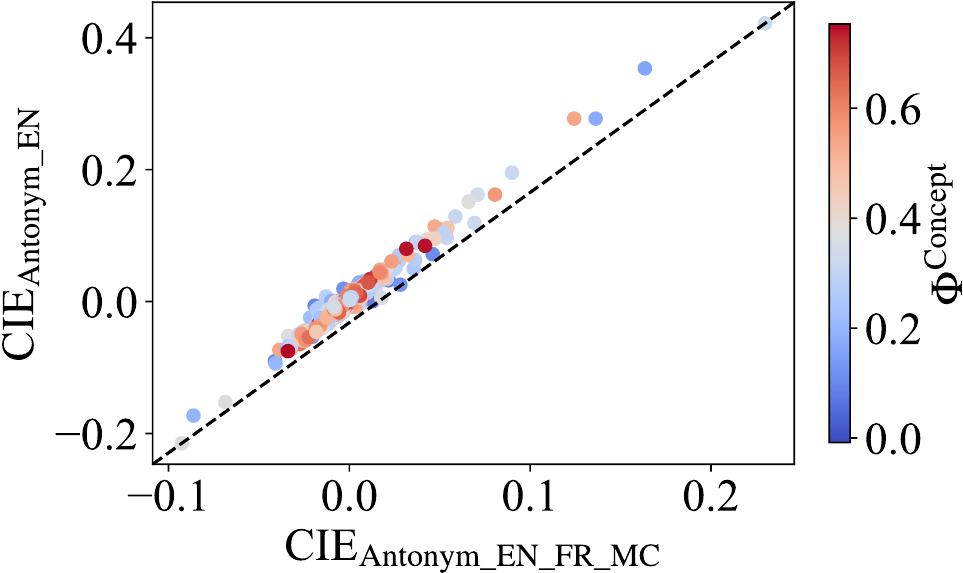}
  \caption{Patching activations from multiple low-level manifestations of a latent concept does not change which attention heads are ranked to have the highest causal effect nor does it help localize latent conceptual information.}
  \label{fig:cie}
\end{figure}

\subsection{Activation Patching Does Not Localize Latent Components} \label{ap_issue}

Attention heads in the \fv s were identified using activation patching on a single low-level manifestation (e.g., English antonyms). To test whether the failure to localize latent conceptual information is due to data selection or the method itself, we computed the CIE for all attention heads for antonyms across three manifestations (\(\text{CIE}_{\text{antonym\_eng\_fr\_mc}}\)) and compared it to \(\text{CIE}_{\text{antonym\_eng}}\).

The top 100 heads ranked by both metrics overlap by 89\%, indicating that adding more low-level datasets does not significantly change the \fv{} composition. 
One might argue that choosing 100 heads is somewhat arbitrary and that varying this number could potentially highlight relational information more effectively. To investigate this possibility, we examined the raw CIE values for each dataset composition. As shown in Figure \ref{fig:cie}, there is a strong correlation between \(\text{CIE}_{\text{antonym\_eng}}\) and \(\text{CIE}_{\text{antonym\_eng\_fr\_mc}}\). In other words, adding more low-level prompts does not alter which attention heads are ranked as having higher causal importance in producing the expected output.

Finally, we note that many attention heads with high \( \Phi^{\text{concept}} \) scores are scored low by the CIE metrics, demonstrating that activation patching is not effective at identifying latent components. More broadly, since activation patching can localize causal, but not latent components, it implies that latent information plays only a small role in next-token prediction (much like knowing an answer to a multiple-choice exam but not the "abcd" response format).

\begin{figure}[h]
  \includegraphics[width=\columnwidth]{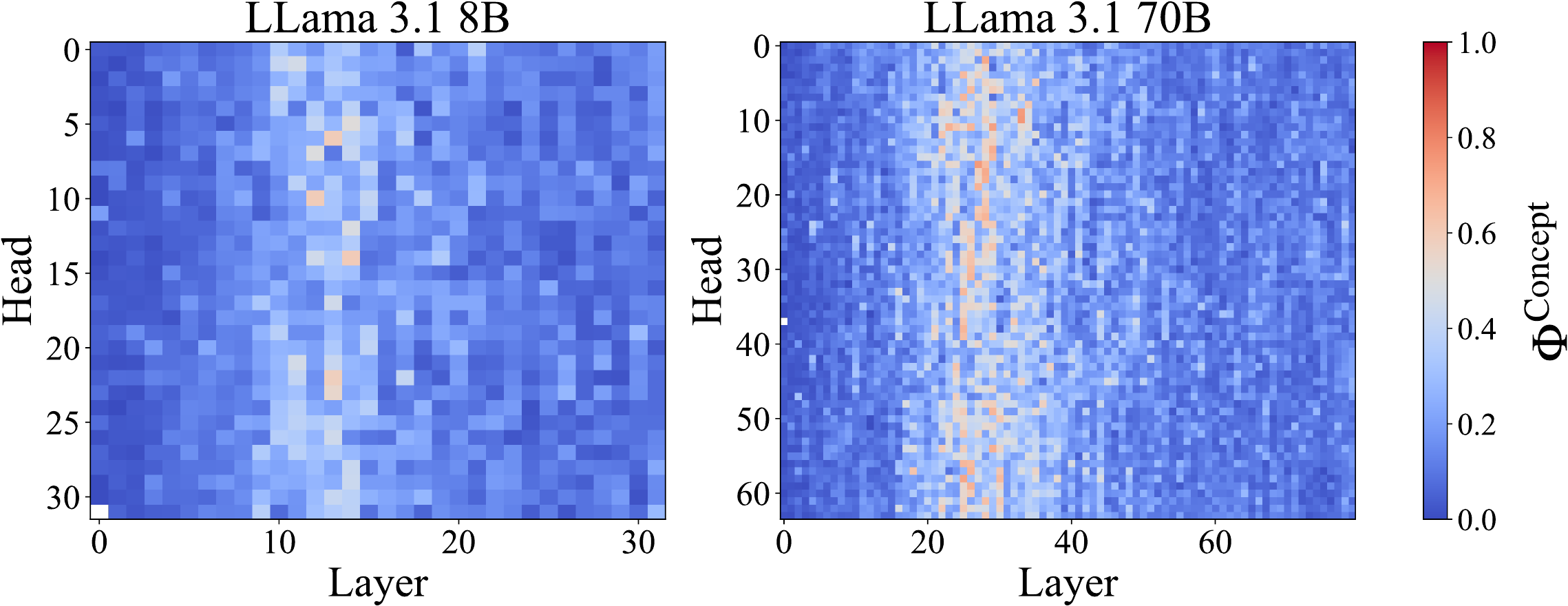}
  \caption{Attention heads encoding verbal concepts emerge in early-to-mid layers.}
  \label{fig:heatmap}
\end{figure}

\section{\cv s emerge for verbal concepts} \label{cv}
In order to distill invariant conceptual representations in LLMs we turn to RSA (Sec. \ref{RSA}). In this section we report on our findings regarding \cv s.

\subsection{\cv s are invariant to low-level transformations} \label{cv_yes_invariance}
Our analysis reveals strong clusters in the \cv{} representational space that are delineated by verbal concepts (Figure \ref{fig:fv_vs_cv}). Compared to the \fv{}s, the \cv{} representations are more \textit{invariant} to low-level transformations and more \textit{specific}—that is, pairwise similarities between different concepts are lower than those within the same concept. While there is a high similarity (Mean = 0.8) among items of the same concept in different languages, the mean similarity drops to 0.7 when items are presented as MC format instead of open-ended. This shows that \cv s, while being close to our notion of conceptual invariance, are not perfect.

\begin{figure}[t]
  \includegraphics[width=\columnwidth]{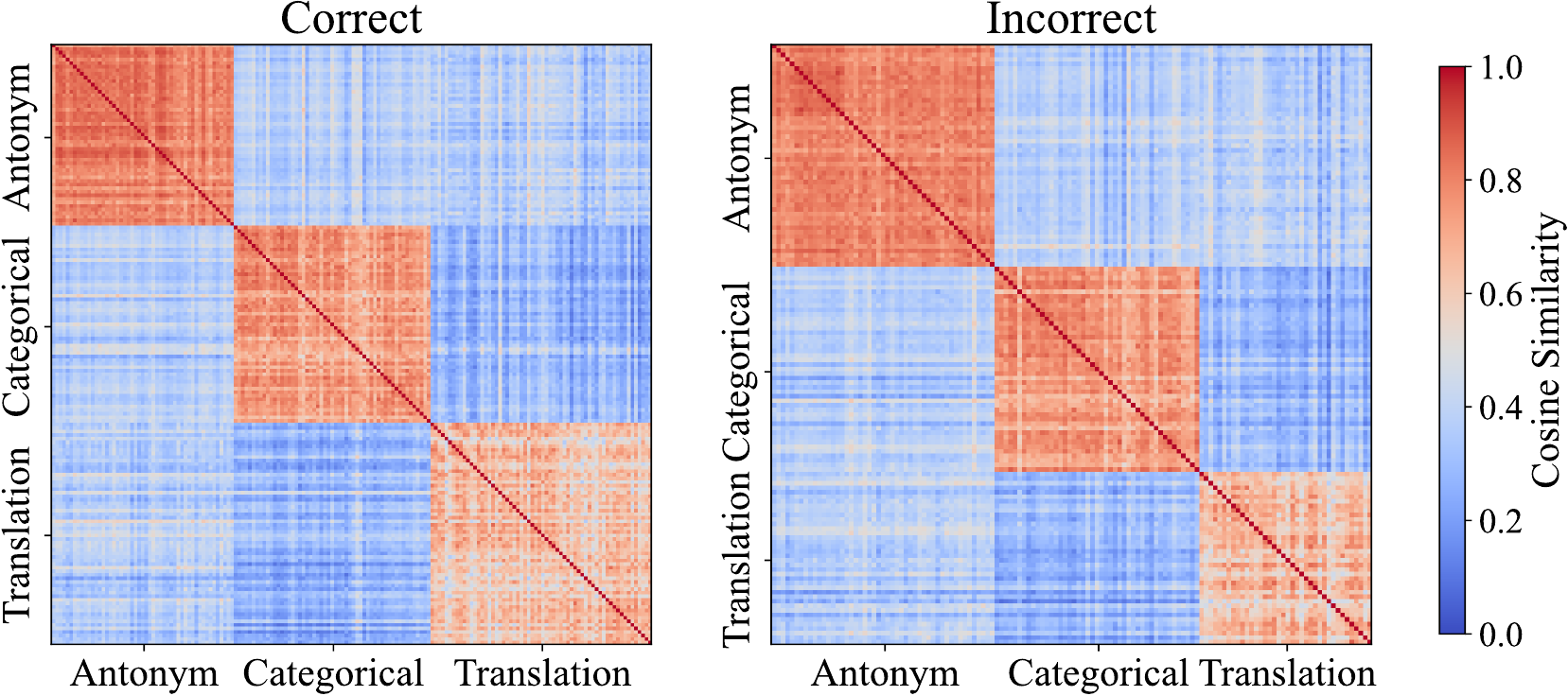}
  \caption{Concept representation can be independent from the model's output. \cv s can encode the correct concept while the model produces the incorrect response. \textit{Note}: we do not show multiple-choice items as performance was too high (\(> 90\%\)) to contrast correct (\(N = \)168) vs incorrect activations (\(N = \)132).}
  \label{fig:acc}
\end{figure}

\subsection{\cv s are feature detectors} \label{cv_feature}

\begin{figure}[t]
  \includegraphics[width=\columnwidth]{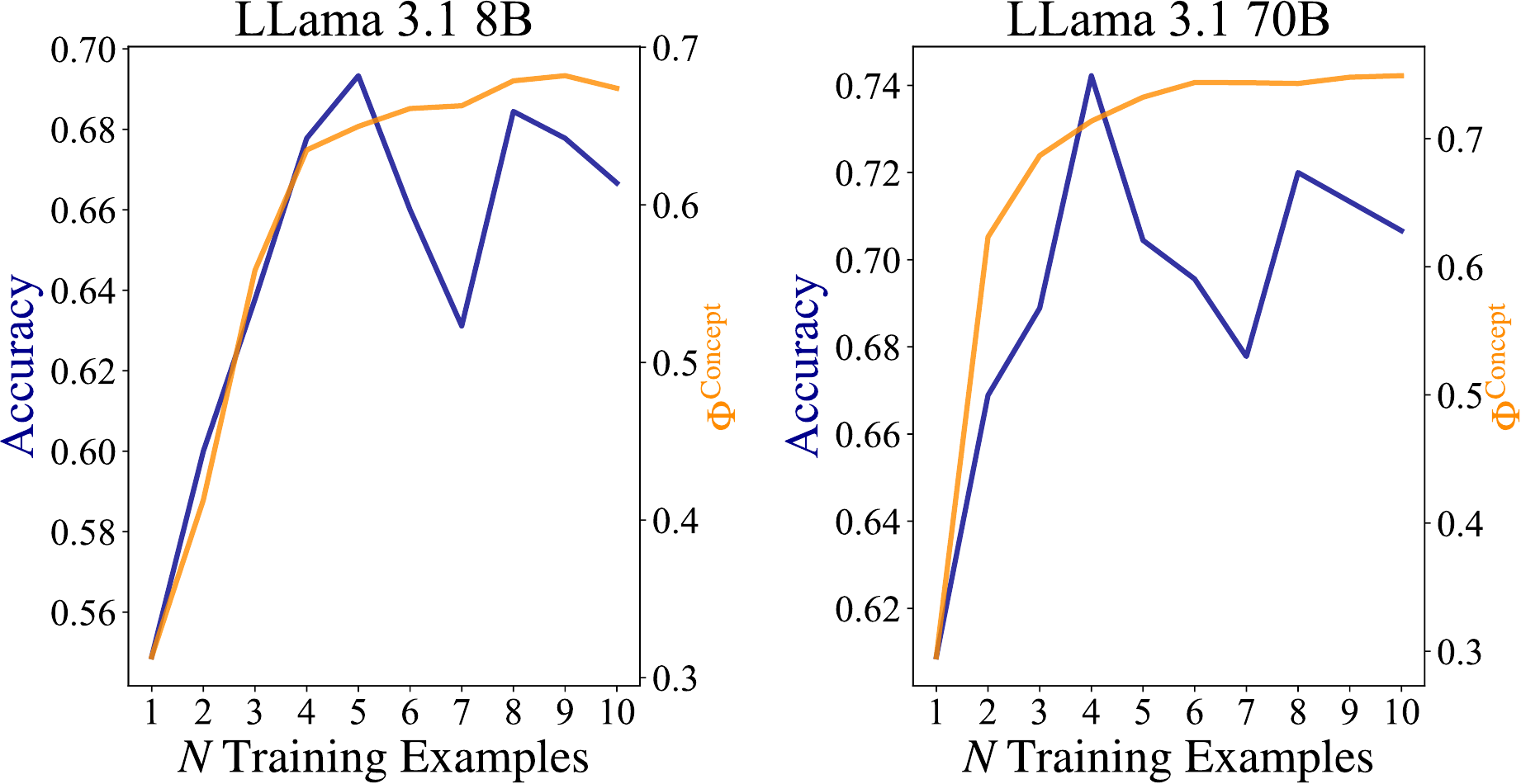}
  \caption{\( \Phi^{\text{concept}} \) grows hand-in-hand with average accuracy over all tasks while \(N\) Training Examples \(< 5\), and then plateaus. \textit{Note}: Error bars around accuracies were removed to reduce clutter.}
  \label{fig:rsa_vs_acc}
\end{figure}

Figure \ref{fig:rsa_vs_acc} shows that model accuracy improves with \( \Phi_{\text{concept}} \) as the number of training examples \(N\) increases, suggesting that the ability to form invariant representations of the underlying concepts is linked to task performance. However, as illustrated in Figure \ref{cv_feature}, the model sometimes forms accurate \cv{}s even when it predicts the incorrect answer. We interpret this as evidence that the model employs \cv{}s as feature detectors. This finding points to a mechanism where the model identifies latent concepts in its early-to-mid layers (see Figure \ref{fig:heatmap}), which may then, or may not, be leveraged in later layers to predict the next token. In cases where the model selects an incorrect token, it may be due either to uncertainty about the specific item or because the correct answer is ambiguous.

\begin{figure}[t]
  \centering
  \includegraphics[width=0.9\columnwidth]{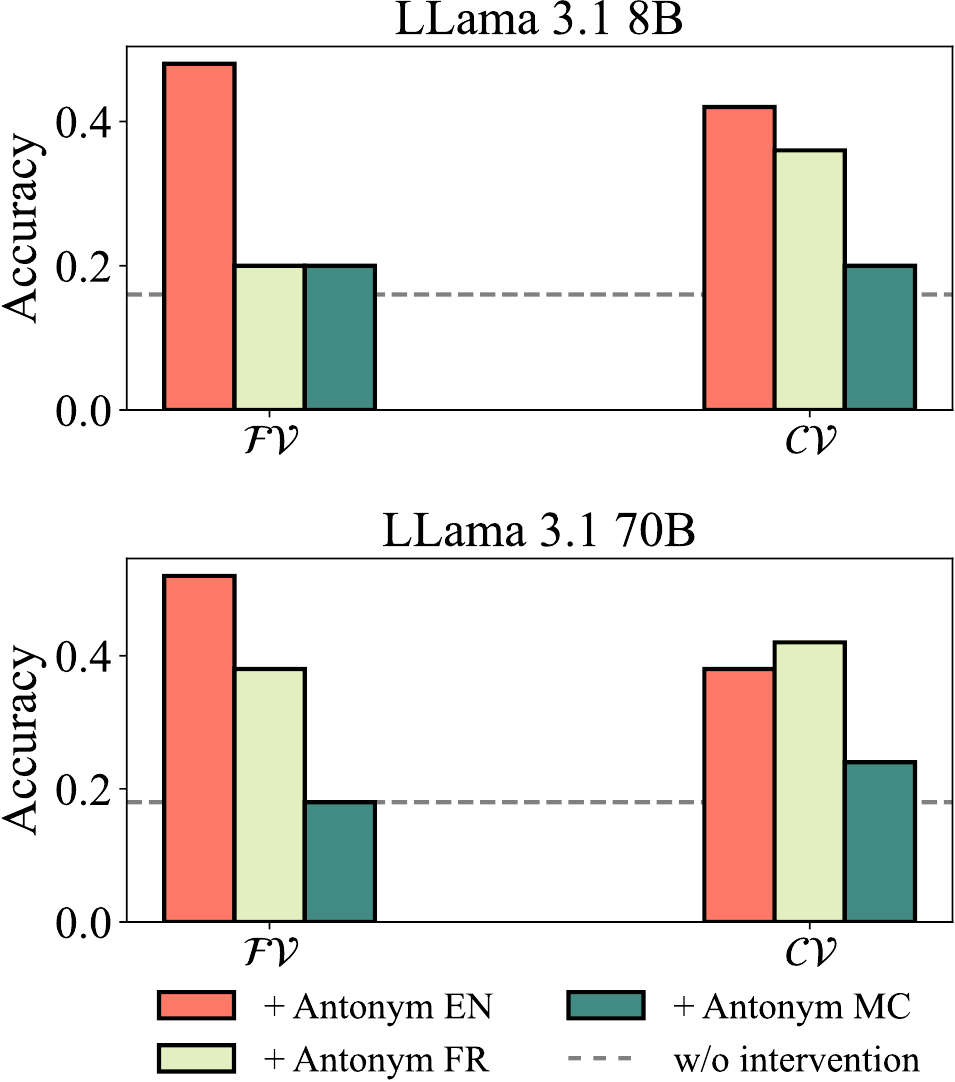}
  \caption{The effect of adding \cv s and \fv s extracted from in-distribution (Antonym EN) and out-of-distribution (Antonym FR and Antonym MC) prompts to the models' hidden states when performing \textit{AmbigousICL}. The grey dashed line shows baseline performance without intervention. \cv s causally guide behaviour model behaviour and are more portable than \fv s.}
  \label{fig:interventions}
\end{figure}

\begin{figure}[t]
  \includegraphics[width=\columnwidth]{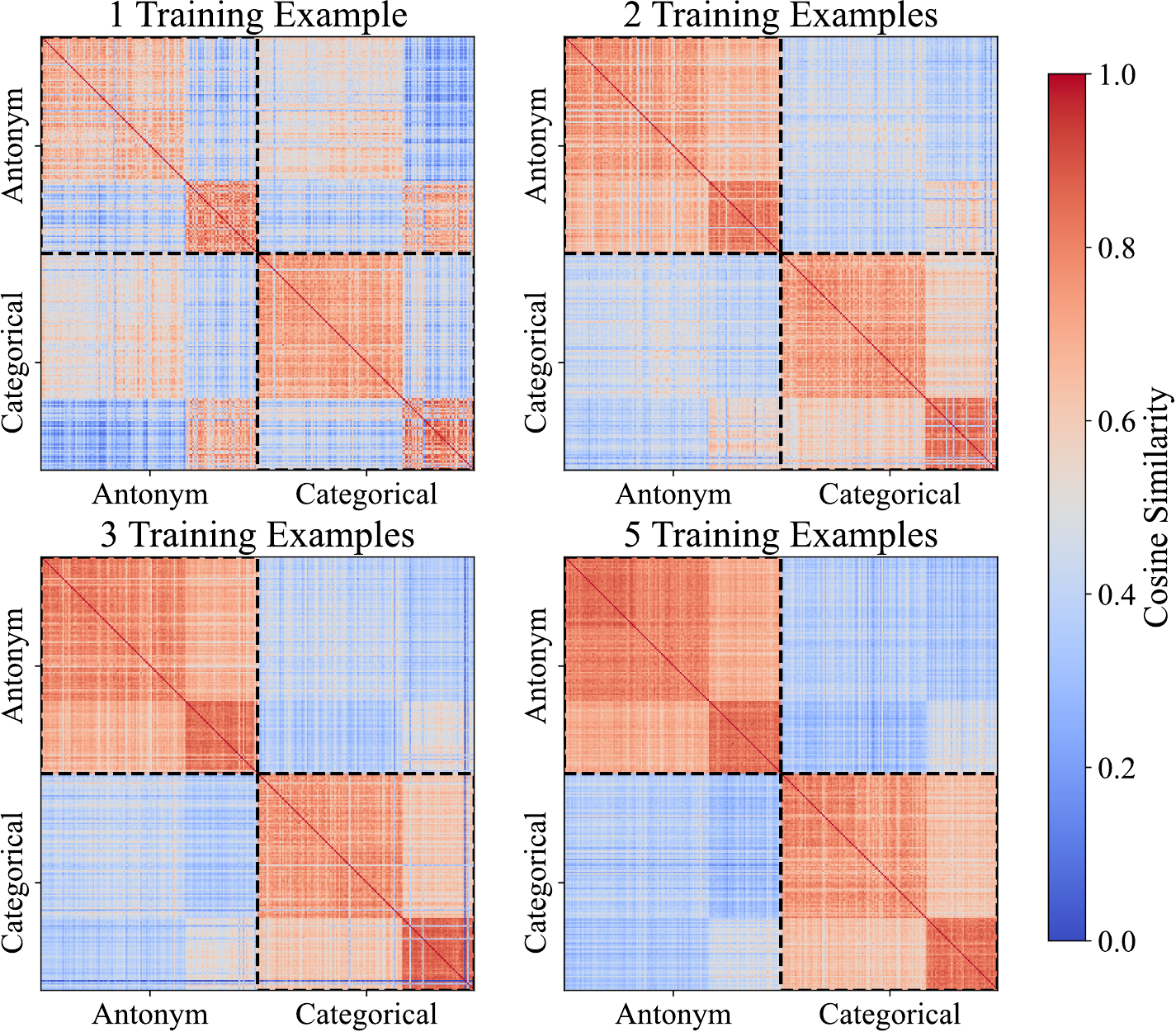}
  \caption{Representational invariance in Llama 3.1 70B grows with the number of training examples in the ICL prompt. The biggest difference is visible from 1 to 2  training examples where \cv s, similarly to \fv s in Figure \ref{fig:fv_vs_cv}), first cluster according to low-level similarity and then display a more invariant representational space, similar to the one in 5 training examples.}
  \label{fig:sim_mats_over_time}
\end{figure}

\subsection{\cv s can causally guide model's behavior} \label{cv_causal}
As we showed, \cv s selectively and invariantly represent verbal concepts, even when the final behavior of the model is incorrect. This raises the question whether the model even uses the information encoded by \cv s. Using causal interventions, and an adapted task we call \textit{AmbigousICL} we show that yes, the models use \cv s.

\subsec{AmbigousICL}
We create a task where we randomly interleave two different ICL concepts in the training examples. 

{
  \centering
  \begin{tcolorbox}[
    width=0.7\linewidth,
    colback=white,
    colframe=black,
    sharp corners,
    boxrule=0.3pt,
    boxsep=2pt,
    left=2pt,
    right=2pt,
    top=2pt,
    bottom=2pt,
    before skip=5pt,
    after skip=5pt
  ]
    {\footnotesize\textbf{AmbiguousICL Example:}}\\[2pt]
    {\small\ttfamily
      \begin{tabular}{@{}l l@{}}
        Q: indoor   & A: \textcolor[HTML]{a53e3a}{outdoor} \\
        Q: noise    & A: \textcolor[HTML]{529bba}{bruit} \\
        Q: western  & A: \textcolor[HTML]{a53e3a}{eastern} \\
        Q: add      & A: \textcolor[HTML]{529bba}{ajouter} \\
        Q: abstract & A: \textcolor[HTML]{529bba}{abstrait} \\
        Q: export   & A:
      \end{tabular}
    }
  \end{tcolorbox}
}

We intervene with \cv s by adding them to hidden states at different layers, \(\mathbf{h}_\ell\), while the model processes a 10-shot \textit{AmbigousICL} prompt and then measure model performance in task execution. We find the best layer to intervene by testing the performance on \textit{AmbigousICL} with the \cv s extracted from 50 prompts in the Antonym EN task. We found these to be layers 14 and 31 for 8B and 70B models respectively (roughly corresponding to where the attention heads encoding verbal concepts emerge, see Figure \ref{fig:heatmap}). For \fv s we follow \citet{toddFunctionVectorsLarge2024} recommendation and use the third of the total layer count. We find that \cv s work best if you apply 10x scaling and 1x for \fv s.

We test both the causal power and the portability of the distilled representations. We extract \fv s and \cv s from three low-level manifestations of the concept Antonym (open-ended EN, open-ended FR, and MC) and transplant them inside of the models while they process the \textit{AmbigousICL} task.

We find that intervening with \cv s increases the probability of model returning the antonym continuation. While \fv s are more effective at guiding the model behaviour when extracted from the same distribution of the task (open-ended EN antonym), they perform worse than \cv s when extracted from Antonym FR (even though \cv s are constructed from a much smaller number of attention heads than \fv s). 

However, when extracting from MC items, performance reduces almost to baseline for both \cv s and \fv s. This provides interesting information regarding how similar vector representations should be in order to achieve similar intervention performance. In case of \cv s the mean similarity of 0.8 between Antonym EN and Antonym FR tasks is enough to achieve the same performance while the similarity of 0.7 between Antonym EN and Antonym MC is not.

Finally, in a zero shot setting \fv s work much better than \cv s ( 50\% vs. 14\% for Llama 8b and 58\% vs. 2\% for Llama 70b). Overall, these results suggest that \cv s capture purer latent conceptual representations, while \fv s also embed lower-level task details that are necessary for correct output.

\begin{figure}[t!]
  \centering
  \includegraphics[width=0.9\columnwidth]{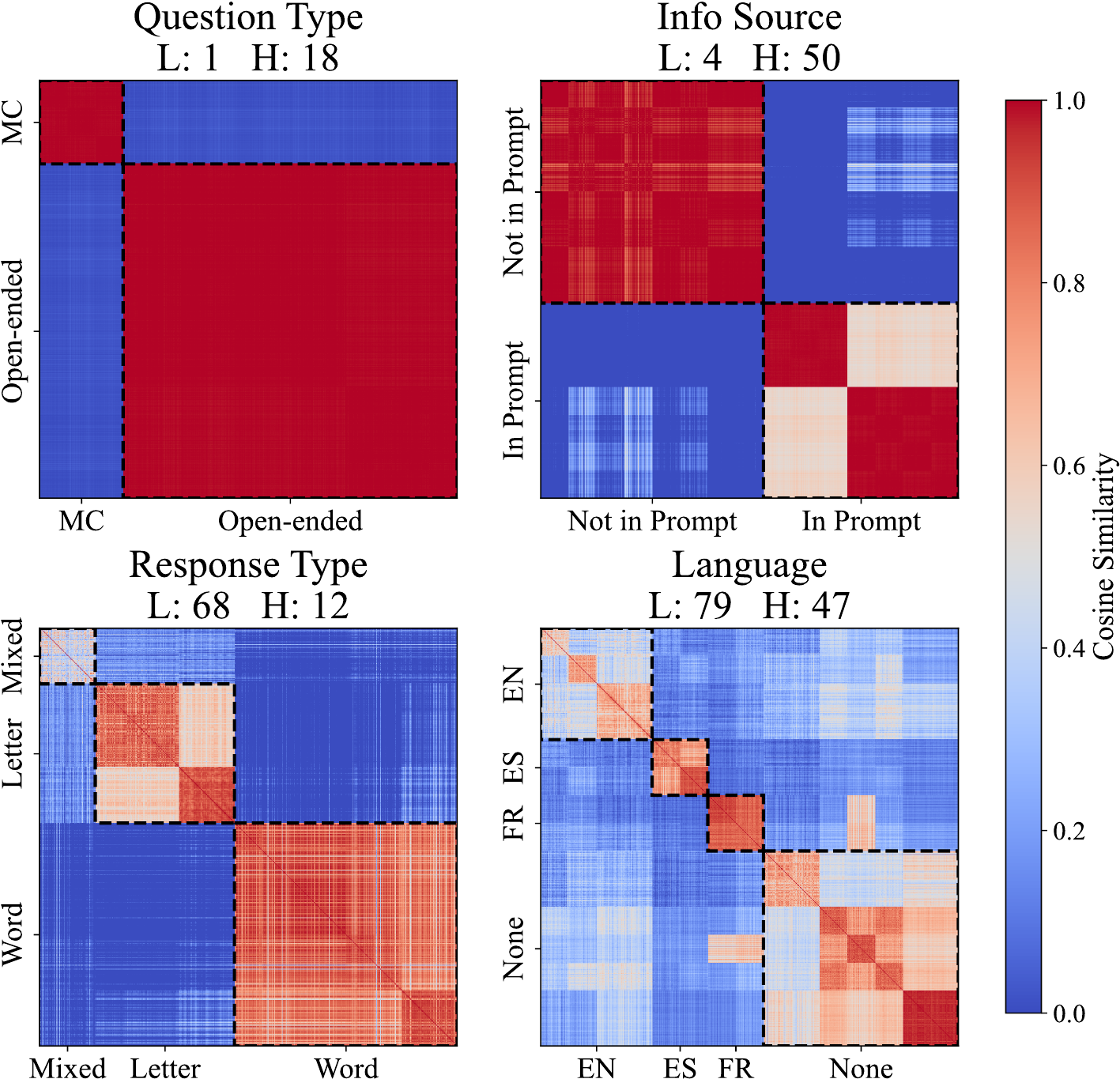}
  \caption{Attention heads with the highest \( \Phi^{q} \) for each task attribute, \(q\). Info source and Question Type emerge early in the transformer, while Language and Response Type in late layers.}
  \label{fig:top_heads}
\end{figure}

\section{Our method also localizes other task attributes} \label{self-aggrandizement}

While this paper focuses on conceptual information in LLMs, we find that using RSA is also fruitful to localize model components where representational spaces align with other task attributes (Figure \ref{fig:top_heads}).

\section{Lack of Abstract Concept Representations Impedes Generalization} \label{generalization}

Figure~\ref{fig:main} shows that abstract concepts are not encoded as linear representations in \cv s. We find no attention heads with \(\Phi^\text{concept\_abstract}\) scores exceeding 0.16 (compared to a maximum \(\Phi^\text{concept\_verbal}\) of 0.75), confirming that abstract representations do not emerge elsewhere in the model. 

However, task performance is high (the 70B model achieves 98\% accuracy for previous/next items and 62\% for abstract previous/next tasks). This implies that LLMs rely on alternative strategies rather than using explicit, top-down representations of abstract concepts such as "Previous" and "Next". One might ask: if the models perform well without abstract representations, what is the drawback? We now show that without reusable abstract concepts, models struggle to generalize to new domains.

\subsec{Letter-string Tasks}\citet{hofstaderGodelEscherBach1979} introduced letter-string analogies to study human analogy-making in a simplified domain. These tasks require understanding “Next” and “Previous” concepts (e.g., given the normal alphabet, if “abc” becomes “abd”, then “ghi” should become “ghj”). \citet{lewisEvaluatingRobustnessAnalogical2024} found that GPT-4’s performance degrades as the alphabet deviates from its canonical order (e.g., “a b c e d f …” is easier than “f e b a d c …”), suggesting that it uses memorization rather than abstraction to solve the task.

We adopt the prompts from \citet{lewisEvaluatingRobustnessAnalogical2024}, extracting \cv s from 20 prompts per alphabet (covering five permuted Latin alphabets and one symbolic alphabet such as “\# \$ * ! @”). Each prompt shows the alphabet with a one-shot ICL example (adapted for non-instruction tuned models). Because Llama 3.1 70B yielded near-zero accuracy on “previous” items, we focus solely on the “next” concept. We also extract \cv s from our "Next Item-in-List" and "Next Abstract-Letter" items (see Sec. \ref{abstract_concepts}).

\begin{table}[ht]
\centering
\small
\begin{tabular}{lcccccc}
\toprule
\(N_\text{perm}\) & 0 & 2 & 5 & 10 & 20 & Symb \\
\midrule
Accuracy & 0.35 & 0.10 & 0.05 & 0.00 & 0.00 & 0.15 \\
\bottomrule
\end{tabular}
\caption{Accuracy in LLama-3.1 70B goes down on Letter-String tasks the more the alphabet deviates from the memorized one (\(N_\text{perm}\)=0). The chance level is 0.04 for the letter alphabets and 0.1 for the symbol alphabet.}
\label{tab:perm_acc}
\end{table}

\begin{figure}[t]
  \includegraphics[width=\columnwidth]{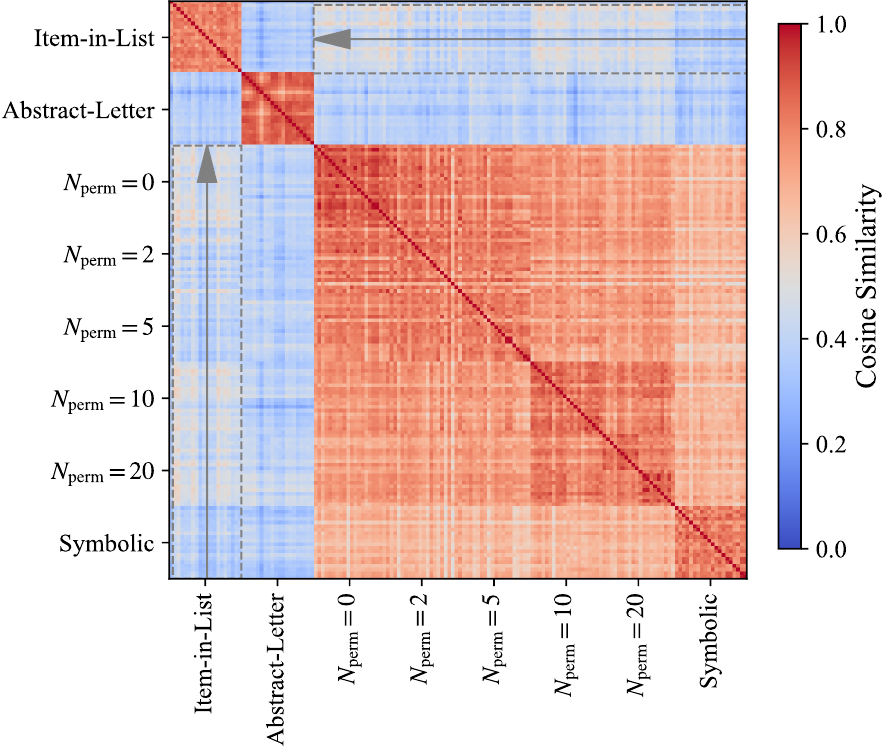}
  \caption{RSM of \cv s extracted from LLama-3.1 70B when performing Letter-String tasks with \(N\) permutations, and other tasks with the concept "Next". The arrows show what the gradient of similarities would look like if the \cv s had a shared representation of ordered lists.}
  \label{fig:generalization_simmat}
\end{figure}

Consistent with our findings so far, we do not see an invariant representation of the concept "Next" across the tasks (Figure \ref{fig:generalization_simmat}). Instead, each task forms its own distinct cluster. Surprisingly, this also suggests that the model represents memorized lists differently in the Next Item-in-List and Letter String tasks. If these representations were shared, we would expect to see a gradient of similarities that decreases with increased alphabet shuffling. This absence might be due to differences between the tasks—for example, the inclusion of the alphabet in the Letter-String prompts or the presence of additional memorized lists in the Next Item-in-List task. In any case, these findings highlight that the model’s representations are highly contextual on these tasks.

\section{Discussion}
We successfully distilled conceptual information from LLM internals for verbal concepts but not for abstract concepts like "previous" and "next". 

Human cognition likely does not process concepts like "next" and "previous" through separate contextual representations. Instead, a shared abstraction—a unified function applied consistently across domains—enables flexible generalization. Investigating whether LLMs exhibit traces of such abstract knowledge, and how to develop it, is critical for achieving human-level artificial reasoning systems.

\section*{Limitations}
A key limitation is our exclusive focus on linear representations (aligned with the Linear Representation Hypothesis \citep{elhageToyModelsSuperposition2022, parkLinearRepresentationHypothesis2024}), despite evidence that LLM representations can be nonlinear \citep{engelsNotAllLanguage2024}. Our LLMs might still encode "Next" and "Previous" nonlinearly but our methods fail to capture it.

Furthermore,  \citet{lampinenLearnedFeatureRepresentations2024} notes that assessing model representations using linear methods can prioritize simpler features, even when complex ones are equally well-learned. Even so, the clear differences between verbal and abstract representations, along with the challenges in abstract tasks, support our conclusion that the "previous" and "next" concepts are either not represented or are represented suboptimally.

Finally, our conclusions are restricted to the LLama-3.1 8B and 70B models, leaving generalizability to other architectures untested.

\bibliography{cv}



\end{document}